# Explainable AI for Chronic Kidney Disease Prediction Using Simulated Federated Learning

Md Zahid Hasan Ontor[1], Md Al Amin[1], Anik Dev Nath[2], and Bikash Kumar Paul[3]

[1] *Dept. of Software Engineering, Daffodil International University, Daffodil Smart City (DSC), Birulia, Savar, Dhaka 1216, Bangladesh*

{zahid35-289, alamin11-593}@diu.edu.bd

[2] *Dept. of Electrical & Electronics Engineering, Ahsanullah University of Science & Technology, Tejgaon, Dhaka, Bangladesh*

anik.eee.aust@gmail.com

[3] *Dept. of Information, Communication & Technology, Mawlana Bhashani Science and Technology University, Santosh, Tangail 1902, Dhaka*

bikash@mbstu.ac.bd

**Abstract.** Chronic Kidney Disease (CKD), characterized by the gradual loss of kidney function, remains a significant public health challenge. Early detection is crucial for preventing severe complications and enhancing patient outcomes. In this study, Federated Learning (FL) with a VotingClassifier was used to predict CKD using a clinical dataset, where Random Forest, AdaBoost, and XGBoost were utilized to compare and identify the best-fitting model for the global server. Additionally, GridSearchCV was applied to optimize the models' performance on the client's side. To enhance model transparency and trustworthiness, explainable AI (XAI) techniques were incorporated to interpret the prediction mechanisms. The global model's average accuracy was 99%, highlighting the potential of interpretable FL models in supporting early CKD diagnosis and advancing data-driven healthcare solutions.

## 1 Introduction

The kidneys are among the most vital organs in the human body. They are responsible for removing waste products and drugs from the body, as well as maintaining the balance of fluids. Additionally, they produce hormones that regulate blood pressure and control the production of red blood cells (RBCs). The process by which kidneys eliminate excess fluid through urine involves highly complex steps that help maintain the body's balance of potassium, acid, and salt. The hormones produced by the kidneys also stimulate RBC production. However, conditions such as diabetes or high blood pressure can lead to kidney disease, which is the primary cause. Individuals with a family history of kidney failure or those who are older are at an increased risk of developing chronic kidney disease (CKD). Symptoms of this disease include persistent fatigue, difficulty concentrating, poor appetite, sleep disturbances, and increased urination, particularly at night. These symptoms are indicative of primary kidney conditions, specifically Acute Kidney Injury (AKI) or Acute Kidney Disease/Disorder (AKD) [1].

Prolonged or untreated AKI or AKD can progress into Chronic Kidney Disease (CKD), a condition in which the kidneys lose their ability to effectively filter waste from the blood or fluids [2]. CKD refers to the gradual deterioration of kidney function, leading to significant difficulties in filtering blood. Individuals with CKD often do not consume adequate water or maintain healthy dietary habits. CKD can also develop in those with type 2 diabetes, high blood pressure, or a family history of kidney disease or failure. Both CKD and Acute Kidney Injury (AKI) can progress to End-Stage Renal Disease (ESRD), a condition where the kidneys lose the ability to function independently [3]. Early identification of CKD can significantly improve a patient's quality of life. This study highlights the need for a reliable prediction model to detect CKD in its early stages. There are many investigations that present the most effective model for predicting CKD at an early stage, achieving optimal performance by utilizing top algorithms with the highest accuracy. The goal is to propose the most suitable and reliable prediction model for CKD while maintaining data privacy.

In this circumstance, Federated Learning (FL) with the integration of Random Forest, AdaBoost, and XGBoost algorithms is effective in addressing the problem while ensuring the privacy of patients' data. FL enables decentralized learning by preventing healthcare institutions from sharing their data with third parties. Here, the Explainable Artificial Intelligence (XAI) analysis approach, specifically LIME, was also applied to interpret and explain model predictions.

The structure of the manuscript is as follows: Section 2 reviews related work, Section 3 outlines the methods and materials, Section 4 presents results and discussion, and Section 5 concludes the study.

## 2 Literature Review

Revathy, S. et al. [4] used Decision Tree, Support Vector Machines (SVMs), and Random Forest algorithms to predict CKD. The results showed that among these algorithms, the Random Forest classifier performed better than the others. In the study conducted by Samatha, K. et al. [5], three machine learning algorithms — Decision Tree, Random Forest, and Logistic Regression — were employed, achieving classification accuracies of 98.75%, 97.5%, and 98.75%, respectively. Although these results demonstrate high performance, there remains potential for further improvement. To enhance model accuracy and generalizability across diverse datasets, the framework should be extended to incorporate additional advanced algorithms capable of achieving superior performance.

Scholar, P. G. et al. [6] focused on predicting Chronic Kidney Disease (CKD) using Naïve Bayes and Decision Tree classifiers. The Decision Tree achieved high performance, with 99.25% accuracy, 99.33% specificity, and 99.20% sensitivity, while Naïve Bayes showed lower accuracy. However, the study was limited to these two algorithms. To improve prediction accuracy and model robustness, exploring a wider range of algorithms is necessary.

In the study by Chittora, P. et al. [7], three feature selection techniques were employed: LASSO regression, the Wrapper method, and correlation-based feature selection. They evaluated seven algorithms, including K-Nearest Neighbors (KNN), Random Forest, Linear Support Vector Machine (LSVM), Logistic Regression, CHAID, C5.0, and Artificial Neural Network (ANN). Among these, LSVM achieved the highest accuracy of 98.86%. The study highlights the need to enhance the model by incorporating additional high-performing algorithms that can deliver greater accuracy and generalize well across various datasets.

Abinaya, U. et al. [8] applied two algorithms (KNN and logistic regression) with accuracies of 95.75% and 98.5%. By systematically evaluating prepared tests and indicators, logistic regression generates a variety of alternative trees. Here, they use only two algorithms, and the other algorithms have been unused. According to the analysis by Rashed-Al-Mahfuz, M. et al. [9], classifiers such as Gradient Boosting (GB), XGBoost (XGB), Logistic Regression (LR), and Support Vector Machine (SVM) were used for CKD prediction. Among these, the Random Forest Classifier outperformed the others in terms of accuracy, achieving scores of 98.25%, 98.50%, 97.25%, and 97.75%, respectively. The study suggests that the prediction model should be further enhanced by incorporating additional algorithms capable of achieving higher accuracy and performing effectively across diverse datasets.

Roy, M. S. et al. [10] investigated the prediction of Chronic Kidney Disease (CKD) using K-Nearest Neighbors (KNN) and Gaussian Naive Bayes. KNN achieved higher accuracy (99.14%) compared to Gaussian Naive Bayes (98.30%). However, the use of only two algorithms limits the model's scope. A more comprehensive approach incorporating a wider range of algorithms is needed to improve prediction accuracy and performance evaluation.

Ekanayake, I. U. et al. [11] utilized two algorithms, Random Forest and Extra Trees Classifier, for CKD prediction, both achieving 100% accuracy compared to other models. These algorithms demonstrated minimal bias toward specific attributes. However, the study emphasized the need to develop a model that incorporates additional algorithms to enhance predictive performance and generalizability.

Almasoud, M. et al. [12] applied classification algorithms to predict Chronic Kidney Disease (CKD), using 10-fold cross-validation for training, testing, and validation. Gradient Boosting (GB) achieved the highest performance, with 99.3% specificity and 98.8% sensitivity. However, only two algorithms were evaluated, limiting the model's potential. Similarly, Ghosh, P. et al. [13] employed four algorithms, with GB achieving the highest accuracy at 99.80% and outperforming others in ROC and AUC metrics. To further improve accuracy and generalizability, future models should incorporate a broader range of algorithms.

# 3 Research Methodology

This study compares Random Forest, XGBoost, and AdaBoost algorithms with the combination of Federated Learning to detect CKD in the initial stage, maintaining proper data privacy. The process begins with transmitting various hyperparameters and client-specific requirements to the applied algorithms. Here the three algorithms play the client's role to make a hybrid model with Federated Learning. Overall, the outline of the method is illustrated in Fig. 1.

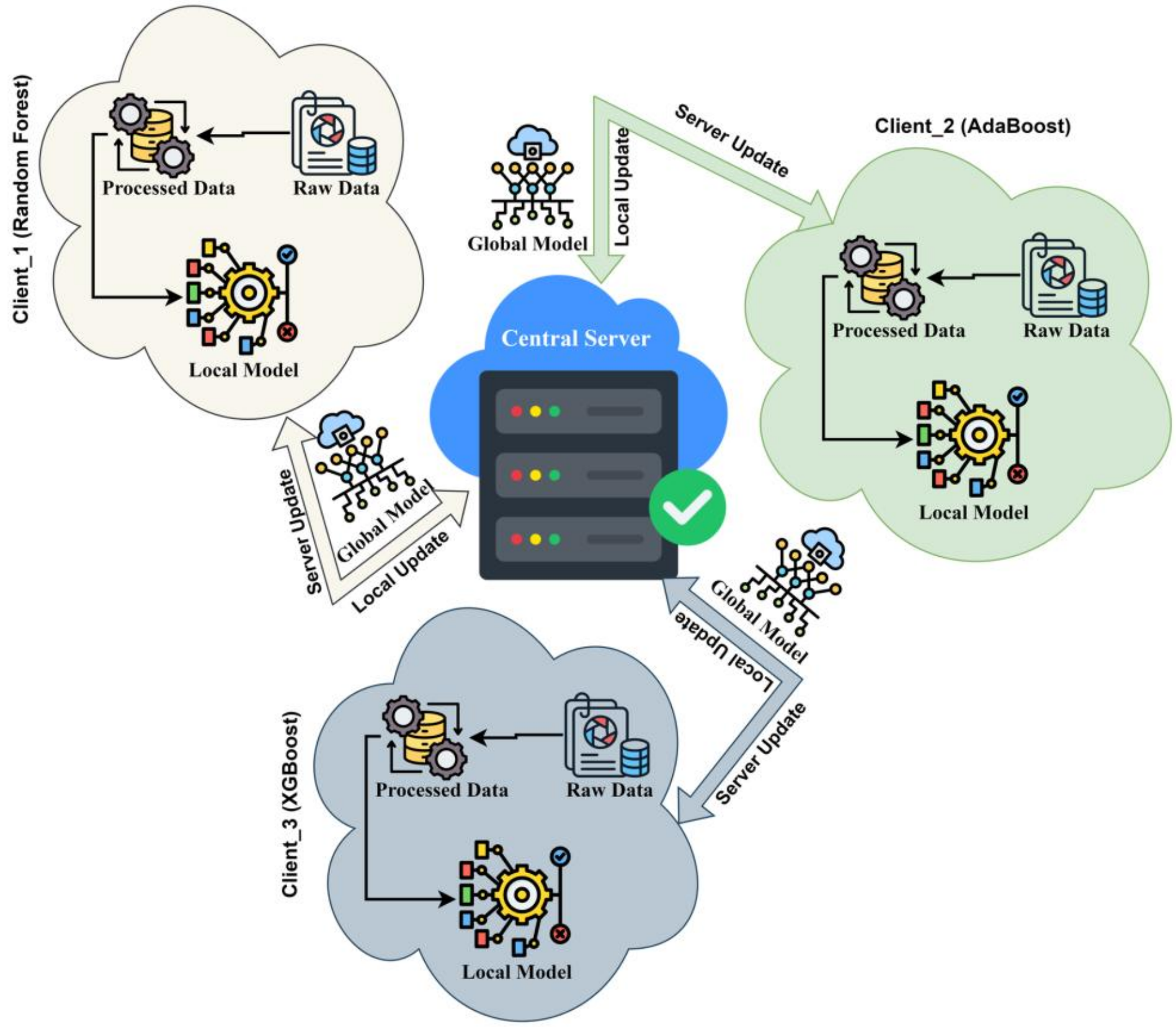

**Fig. 1.** Federated learning architecture with three clients collaboratively training an ensemble model (Random Forest, XGBoost, AdaBoost) without sharing raw data. Each client trains local models on private data, which are sent to a central server for aggregation and used to update a global model, thereby preserving data privacy.

In the first step, manipulate the selected dataset. To reduce noise and irregularities and balance the class distribution, allowing the model to learn more effectively from all classes, the SMOTE technique was applied. Then the three clients train the data with the touch of the GridSearchCV optimizer, where the optimizer automates the process of finding an optimal set of parameters and helps avoid overfitting. Also, it finds the optimal parameter values from the local models and loss function. LIME analysis is used for interpreting the behavior of algorithms. After that, the updated model parameters are frequently sent back to the federated learning server.

Fig. 1 illustrates the comparative performance of classification models across three federated clients. For Client 1, the Random Forest algorithm achieved the highest performance across evaluation metrics. In Client 2, AdaBoost outperformed other models, while in Client 3, XGBoost demonstrated superior predictive capability. Then the server aggregates the model updates from all participating clients. The procedure continues until satisfactory results are achieved or a consistent reduction in loss is observed. Ultimately, the global results are combined with the outcomes from each client to evaluate the overall performance.

### 3.1 Dataset Description

In this experiment, a CKD dataset obtained from Kaggle [14] was used to develop and evaluate the prediction model. The dataset contains 400 data points with 14 medical predictor variables. Among these, the class attribute is the sole dependent variable, while the remaining parameters serve as independent variables. The class label indicates whether a patient has CKD or not. According to the dataset distribution, 62.5% of the patients are diagnosed with CKD, while 37.5% are not. Features of the dataset are illustrated in Table 1.

**Table 1.** Dataset Features Description

| PD | Type | Description |
|---|---|---|
| Class | Binary | Patient's condition |
| Bp (Blood Pressure) | Numerical | Human min Bp value is 50 and the max value is 180 |
| Sg (Specific Gravity) | Numerical | Density of patients' urine compared to the water |
| Al (Albumin) | Numerical | Normal level is 3.4–5.4 g/dl. Lower means malnutrition and higher means acute infection [15] |
| Su (Sugar) | Numerical | Normal blood sugar 140 mg/dl is normal |
| Rbc (Red Blood Cell) | Numerical | Number of Rbc in patient's blood |
| Bu (Blood Urea) | Numerical | Wastes of patient's blood |
| Sc (Serum Creatinine) | Numerical | Patient's blood creatinine amount |
| Sod (Sodium) | Numerical | The normal sodium level in the blood ranges from 135 to 145 mEq/L. When the sodium level drops below this range, hyponatremia occurs |
| Pot (Potassium) | Numerical | Normal potassium level 3.6–5.2 mmol/L in blood. If Pot > 6 then dangerous |
| Hemo (Hemoglobin) | Numerical | Amount of RBC. If Hemo level is low, then it's called anemia [16] |
| Wbcc (White Blood Cell Count) | Numerical | Wbcc count in the patient's body. It produces more WBC to battle and combat bacteria, pathogens, etc. [17] |
| Rbcc (Red Blood Cell Count) | Numerical | Number of Rbcc in patients' blood. Normal is 4.2–5.6 (Women) and 4.7–6.1 (Men) [18] |
| Htn (Hypertension) | Numerical | Blood pressure is measured by systolic and diastolic values. Htn occurs when systolic pressure is above 120 mmHg and diastolic is below 80 mmHg [19] |

## 3.2 Dataset Preprocessing

The dataset consists of information from 400 CKD patients and includes 14 parameters. The class column serves as the dependent variable, while the remaining columns are independent variables that have a strong correlation with the class column. Initially, the features were separated from the target variable. In Fig. 2, the pie chart illustrates the relative frequency of CKD and non-CKD instances in the dataset. Here, 1 represents CKD and 0 represents not CKD. Among the 400 data points there were 250 (62.5%) for CKD and the rest 150 (37.5%) for not CKD.

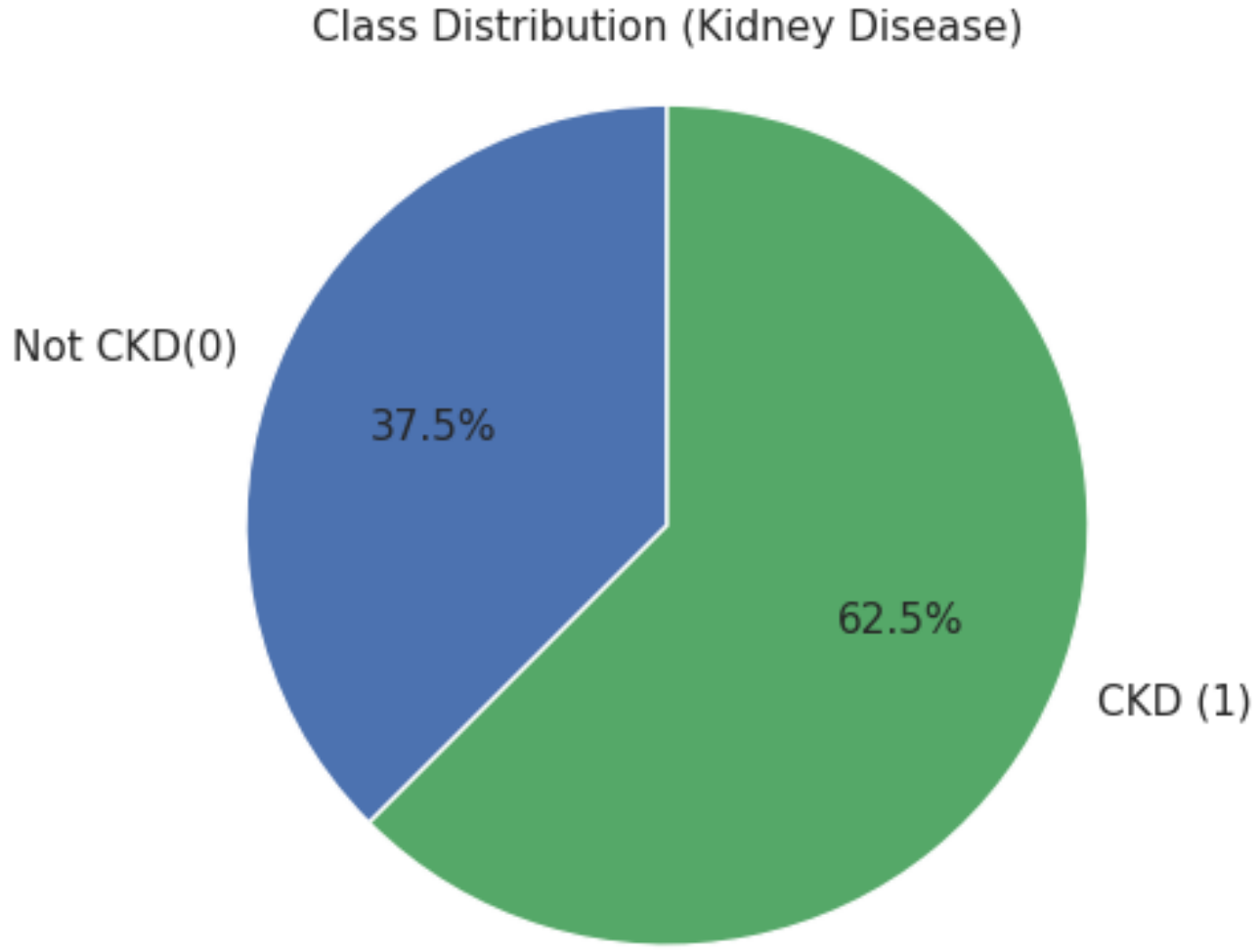

**Fig. 2.** Chronic Kidney Disease (CKD) Class Distribution Overview.

To address class imbalance, the Synthetic Minority Over-sampling Technique (SMOTE) was applied to generate synthetic samples for the minority class. After applying this process, normal patient data soared to 250 from 150. On the other hand, chronic kidney patient data remains unchanged, and for this reason, the dataset increased to 500 from 400. The resulting balanced dataset was then partitioned into three distinct subsets, each simulating a client in a federated learning environment. This partitioning was performed using stratified splitting to ensure that the original class distribution was preserved within each client's dataset. Subsequently, each client's data was further stratified into training and testing sets. To avoid potential model fitting issues, clients whose training data lacked representation from both classes were excluded from further processing.

### 3.3 Ensemble Classification Algorithm

Ensemble methods enhance model accuracy by combining multiple models instead of relying on a single one. This approach significantly boosts the precision of predictions. By integrating various methods, ensemble techniques are particularly effective in improving predictability, especially in classification tasks, where they help reduce bias and increase accuracy.

- **Random Forest (RF)** is an advanced and effective ensemble learning technique widely recognized for its classification capabilities. It consists of multiple decision trees, each trained on different feature subsets of the training data. When a new test sample is introduced, each tree in the forest independently evaluates it and generates a classification outcome. The RF algorithm then determines the predicted class of the test data by taking a majority vote across all the trees in the network.
- **XGBoost** is a highly effective machine learning algorithm that has gained significant popularity, particularly for its outstanding performance in Kaggle competitions involving structured datasets. It is known for its high-speed execution and exceptional model performance, making it a preferred choice for many data scientists. XGBoost is an efficient and versatile implementation of gradient-boosted decision trees, optimized for speed, flexibility, and portability [20]. By leveraging the Gradient Boosting framework, XGBoost efficiently builds machine learning models using parallel tree boosting, allowing it to address a wide range of data science challenges both quickly and accurately.

- **AdaBoost** is a boosting technique used in machine learning as an ensemble method. It works by adjusting the weights of instances, assigning higher weights to those that are incorrectly classified, a process known as Adaptive Boosting [21]. This method enhances classification by converting weak learners or predictors into stronger ones. However, AdaBoost has some limitations, including its reliance on objective data, which makes it particularly vulnerable to uniform noise. Additionally, if the weak classifiers are too weak, they can lead to low margins and increase the risk of overfitting.

## 3.4 Processed Federated Model

Federated Learning (FL) is a distributed machine-learning technique that involves a central server and a group of clients. Computing nodes known as clients use their local data to conduct local training. For each client, the model achieving the highest accuracy on its respective test set was selected as the optimal local model. These client-specific best models were subsequently integrated into a global model using a weighted VotingClassifier. The weights assigned to each model within the ensemble were proportional to their respective test accuracies, thereby granting greater influence to models with higher performance. The resulting global model was then trained on the aggregated training data from all clients and evaluated on a combined test set to assess its overall performance. Many real-world applications with stringent privacy requirements can benefit greatly from its special distributed training mode and security aggregation approach [24].

It is a collaborative machine learning approach that utilizes locally available data samples to train an algorithm on several servers or devices without ever sharing the actual data. Local computing and model transfer are strongly maintained by FL, which lowers some of the costs and systemic privacy problems associated with standard central machine learning techniques. The client's original data is kept locally and cannot be moved or shared [25]. At the same time, it can scale into a larger dataset by spreading across a greater number of machines or devices, efficiently handles the challenge of bandwidth consumption, several devices work together without exchanging data, and model parameters are transmitted to a central server to strengthen the model while maintaining solid data privacy concerns.

## 3.5 Explainable Artificial Intelligence (XAI)

Healthcare is a highly sensitive domain where transparency plays a critical role in maintaining public trust. People are more likely to trust AI systems when they can understand the underlying mechanisms behind the model's predictions. Explainable AI (XAI) addresses this need by offering insights into both the predictions and the rationale behind machine learning model decisions [22]. Among various XAI methods, LIME (Local Interpretable Model-Agnostic Explanations) has emerged as a powerful and widely used approach for interpreting the behavior of algorithms [23], providing clear and consistent explanations of individual and global model outputs.

LIME is a powerful game-theoretic approach used to interpret machine learning model predictions by assigning each feature a LIME value that quantifies its contribution to the final output. It ensures a fair, consistent, and additive distribution of model output, enabling both local (instance-level) and global (model-wide) explanations. One of LIME's strengths lies in its ability to make comparative analyses across different models and predictions due to its solid theoretical foundation. Furthermore, although LIME is considered model-agnostic in theory, the practical implementation of efficient and scalable algorithms varies significantly across model types, and optimized solutions are not always readily available for every architecture.

### 3.6 Performance Evaluation Metrics

A simulated federated learning model was assessed using key evaluation metrics derived from the confusion matrix. The confusion matrix included counts of True Positives (TP), False Positives (FP), True Negatives (TN), and False Negatives (FN), based on comparisons between actual and predicted data. These counts were then used to calculate various performance metrics for evaluating the model.

In this study, several key evaluation metrics are derived from the confusion matrix using their respective formulas. All formulas describe the important relationship between the actual class and the predicted class. Here, all employed models were evaluated and compared using the following formulas.

Accuracy is a fundamental performance metric that quantifies the proportion of correctly classified instances among the total number of samples. In the context of chronic kidney disease (CKD) prediction, accuracy refers to the number of correct CKD predictions divided by the total number of instances in the dataset. Accordingly, accuracy is computed using Equation 1.

$$Accuracy = (TP + TN) / (TP + TN + FP + FN) \qquad (1)$$

Recall is a performance metric that measures a model's ability to correctly identify patients who truly have Chronic Kidney Disease. It represents the proportion of actual CKD cases that the model successfully detects. Hence, Recall is calculated according to Equation 2.

$$Recall = TP / (TP + FN) \qquad (2)$$

Precision refers to the proportion of predicted CKD cases that are true CKD cases. It helps assess how reliable the model's positive predictions are. Equation 3 mathematically represents precision.

$$Precision = TP / (TP + FP) \qquad (3)$$

F1-Score provides a balanced metric by combining both precision and recall through their harmonic mean. It is critical to detect as many true CKD cases as possible while also maintaining a low false alarm rate. Equation 4 is used in the computation of the F1-Score.

$$F1\text{-}Score = 2 \times (Recall \times Precision) / (Recall + Precision) \qquad (4)$$

## 4 Result and Discussion

### 4.1 Client-Specific Best Model Accuracy

This section outlines the analytical procedures designed to determine the most effective approach for predicting CKD. Here, the three applied algorithms from three different clients were trained with the balanced dataset to find the output. Fig. 3 displays the performance measures (Accuracy, Recall, Precision, and F1-Score) of the best-performing machine learning model for each of the three simulated clients in the federated learning setup. The x-axis represents the three clients and the y-axis indicates the score for each metric, ranging from 90% to 100% for better visualization of high scores. The plot shows that for Client 1, all four metrics achieved a score of 100%. For Client 2 and Client 3, Accuracy, Recall, Precision, and F1-Score are approximately 97%, 100%, 94%, and 97%, respectively. This visualization allows for a direct comparison of the local model performance across different clients before aggregation, highlighting potential variations in model effectiveness on client-specific data subsets.

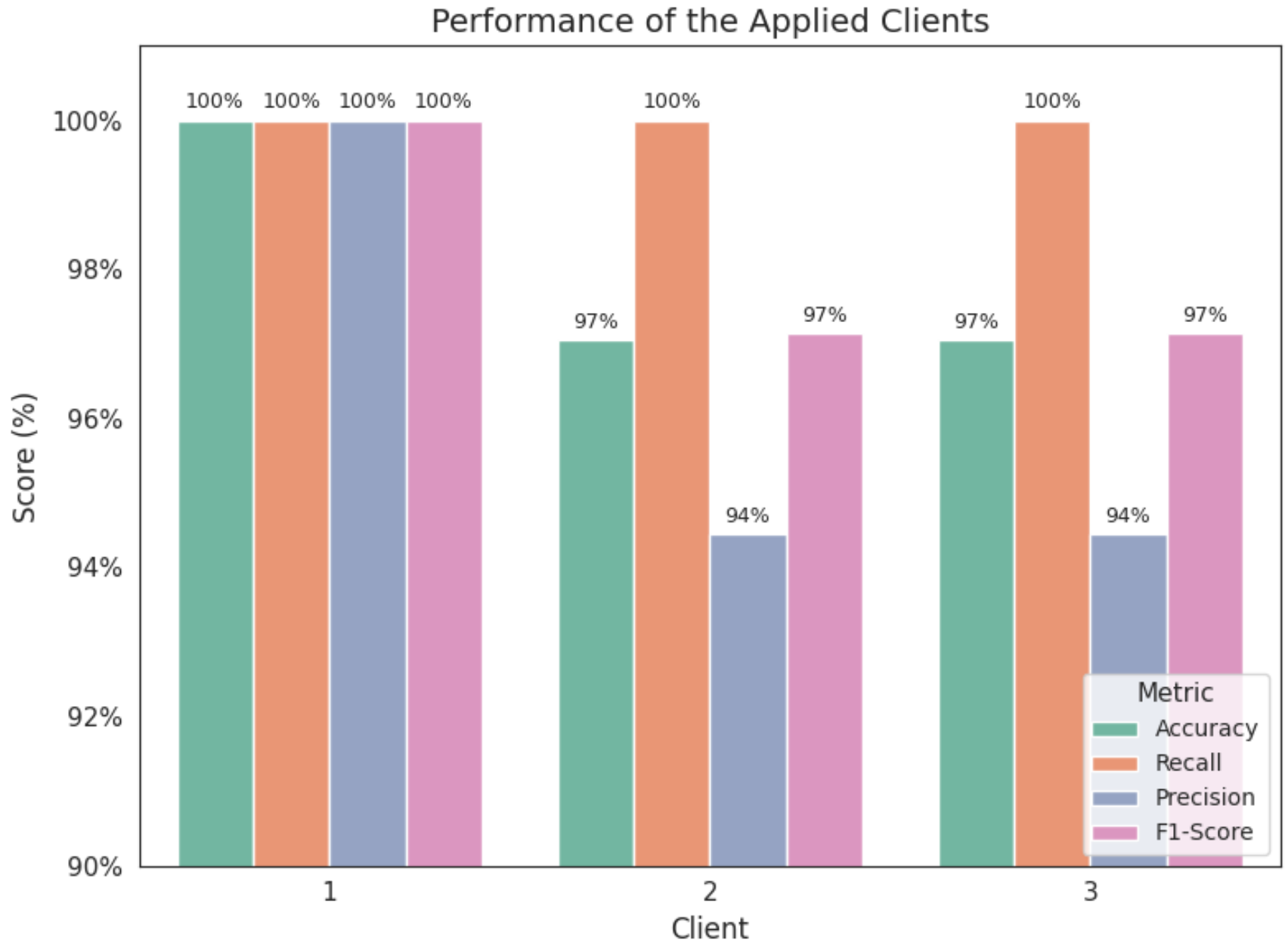

**Fig. 3.** Comparison of Best Performing Local Model Accuracy, Recall, Precision, and F1-Score Across Three Different Clients.

## 4.2 Global Model Aggregation and Evaluation

Here the best models are aggregated into a global model using a weighted VotingClassifier, where weights are based on client model accuracies. It combines the predictions of multiple individual models and uses the predicted probabilities from each constituent model and averages them to make a final prediction.

The classification of the global model's performance metrics for the two classes (Class 0 and Class 1) is shown in Table 2. For Class 0, the model's Precision and Recall were 98% and 100%, respectively, and for Class 1, they were 100% and 98%, respectively. For both classes, the F1-score, which balances recall and precision, is continuously high at 99%. Additionally, the model's accuracy for both classes is 99%, demonstrating a robust and well-rounded prediction performance across the target classes.

**Table 2.** Global Model Evaluation

| Precision | Recall | F1-score | Accuracy | Class |
|---|---|---|---|---|
| 98 | 100 | 99 | 99 | 0 |
| 100 | 98 | 99 | 99 | 1 |

The global model was evaluated using 5-fold cross-validation, visualizing the accuracy scores for each fold. The performance of the global federated learning model over five-fold cross-validation is depicted in Fig. 4 as a bar plot. Every single fold in the cross-validation is represented on the x-axis (designated “Fold 1” through “Fold 5”). When tested on the test set associated with that particular fold, the model's accuracy score is displayed on the y-axis. For a specific fold, the accuracy is shown by each bar. The plot shows that the model performs consistently throughout the dataset's various partitions, with mean cross-validation accuracies ranging from 96% to 100%, with the highest performance observed in Folds 3 and 5 (100%), and the lowest in Fold 2 (96%). This research sheds light on the trained global model's generalization capacity and resilience.

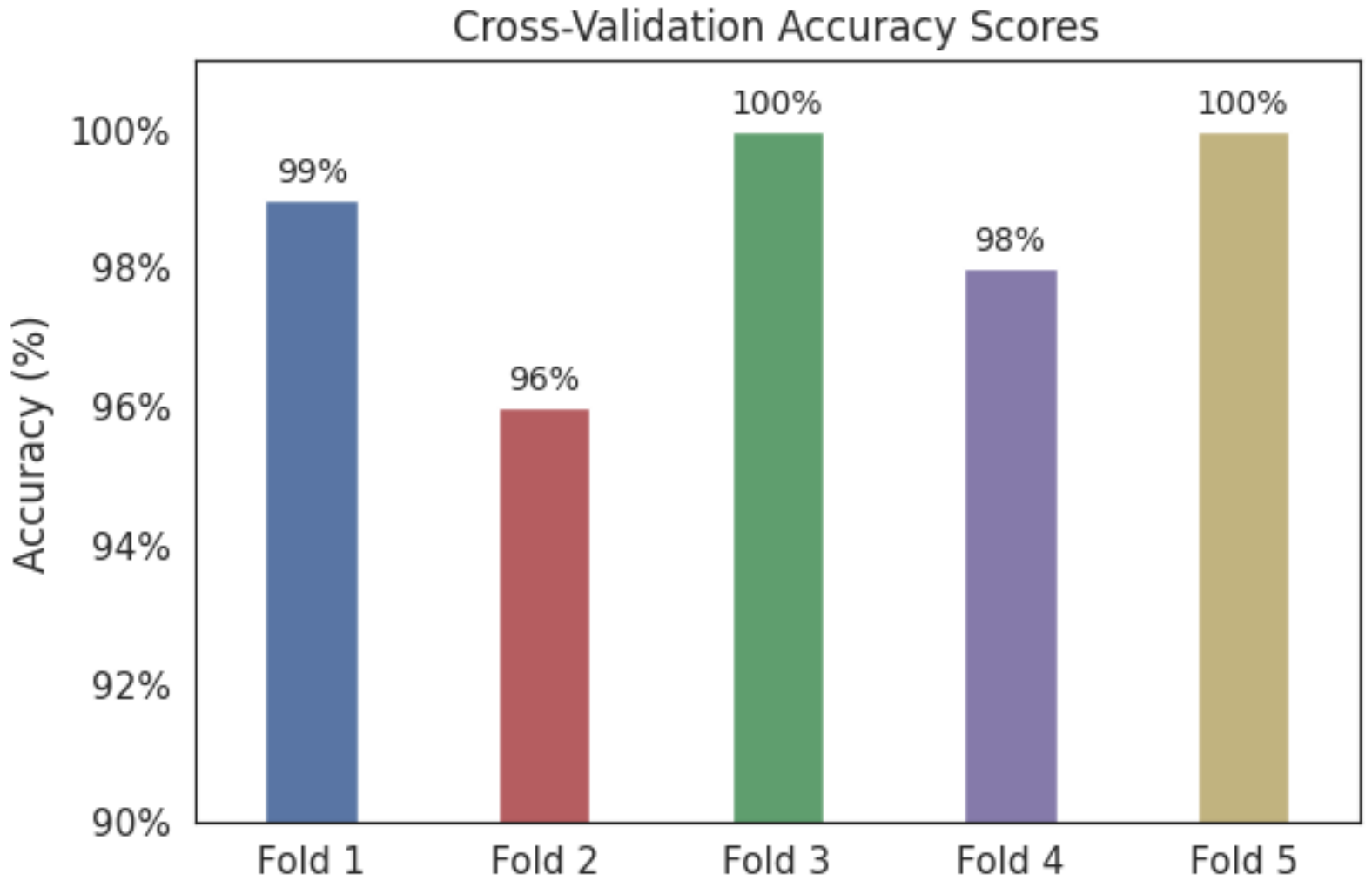

**Fig. 4.** Cross Validation Accuracy Scores for the Global Federated Learning Model.

Explainable AI gives a sense of trust to the AI application user that can provide insight into the model prediction mechanism and feature contribution.

Fig. 5 shows the feature-level contributions for a single test instance in the context of renal disease prediction using a Local Interpretable Model-agnostic Explanations (LIME) graphic. Along the y-axis of the chart are a number of input feature intervals, each of which is connected to a corresponding weight on the x-axis that shows how it affected the model's choice. Features that support the projected class are shown by positive weights (in green), whereas features that have opposing effects are indicated by negative weights (in red). Interestingly, the "$11.30 < \text{Hemo} \leq 13.23$" interval shows the highest positive contribution, increasing confidence in the expected result, while the "$\text{Htn} \leq 0.00$" interval shows the strongest negative influence, diminishing the prediction.

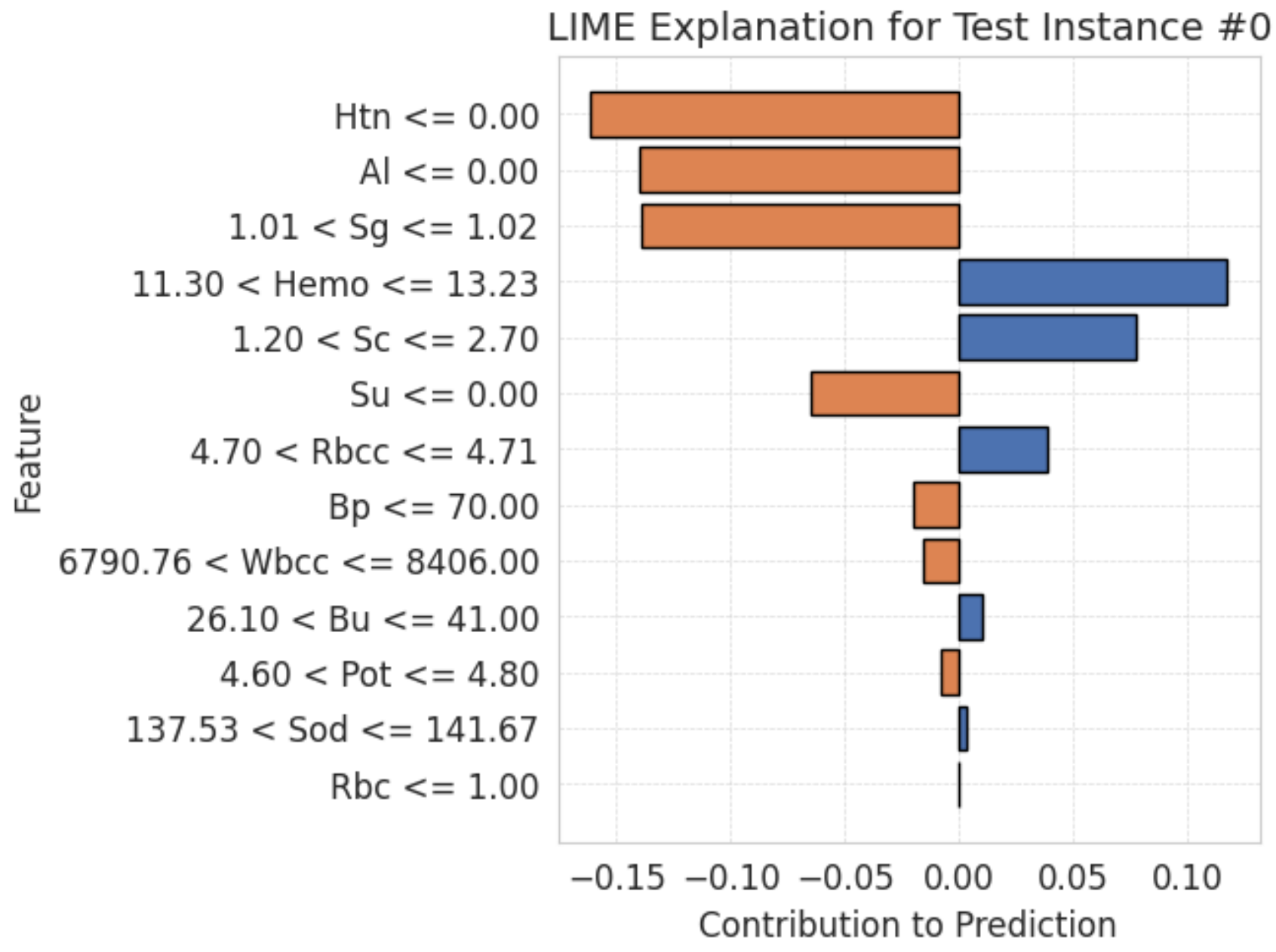

**Fig. 5.** Feature Contribution Visualization Using LIME for Kidney Disease Prediction.

## 5 Conclusion

This study demonstrates the effectiveness of ensemble machine learning models (Random Forest, AdaBoost, and XGBoost) integrated with Federated Learning, which preserves clients' privacy. First, SMOTE was applied for data balancing, then GridSearchCV for finding the optimal parameter values from each method. After that, the VotingClassifier was integrated with Federated Learning to add weights from the clients. Finally, XAI (LIME) techniques are used to enhance the transparency of the predictive system. The model not only delivered high predictive performance but also provided interpretability, an essential factor in gaining trust from healthcare professionals regarding its privacy concerns. Among the features analyzed, hemoglobin and sugar levels emerged as key indicators for CKD prediction. The exceptional average accuracy of the global model obtained 99%, highlighting the potential of FL-based solutions in supporting early diagnosis and reducing dependency on extensive clinical testing. Future work will focus on validating these findings across larger, more diverse datasets to enhance clinical applicability, ultimately contributing to more effective CKD management.